\pdfoutput=1

\documentclass[11pt]{article}

\usepackage[final]{acl}

\usepackage{times}
\usepackage{latexsym}

\usepackage[para]{threeparttable}
\usepackage{booktabs}
\usepackage{multirow}
\usepackage{amsmath}
\usepackage{amssymb}
\usepackage{xcolor}
\usepackage{hyperref}
\usepackage{mathtools}
\usepackage[ruled,vlined]{algorithm2e}
\usepackage{tcolorbox}
\usepackage{listings}
\usepackage{enumitem}
\usepackage{makecell}
\definecolor{c1}{HTML}{0049C0}
\usepackage{colortbl}
\usepackage{tabularx}
\definecolor{mygray}{gray}{.95}
\definecolor{mycell}{rgb}{0.85, 0.93, 0.97}
\definecolor{mycelltwo}{RGB}{255, 238, 241}
\definecolor{Ground}{RGB}{255,184,55}
\definecolor{Rice}{RGB}{251,248,238}
\definecolor{Dirt}{RGB}{191,169,115}
\definecolor{Pink}{RGB}{226,184,176}
\definecolor{Violet}{RGB}{163,148,170}
\definecolor{mygray}{RGB}{226, 226, 226}

\usepackage{longtable}
\usepackage{arydshln}
\tcbuselibrary{breakable}
\usepackage{amssymb}
\makeatletter
\renewcommand{\maketag@@@}[1]{\hbox{\m@th\normalsize\normalfont#1}}%
\makeatother

\newcommand{\fname}{\textsc{CHIQ}}

\newcolumntype{g}{>{\columncolor{Ground!10}}c}
\newcolumntype{d}{>{\columncolor{Dirt!10}}c}
\newcolumntype{f}{>{\columncolor{Pink!10}}c}
\newcolumntype{v}{>{\columncolor{Violet!10}}c}
\newcolumntype{P}[1]{>{\centering\arraybackslash}p{#1}}

\usepackage[T1]{fontenc}

\usepackage[utf8]{inputenc}

\usepackage{microtype}

\usepackage{inconsolata}

\usepackage{graphicx}

\definecolor{amaranth}{rgb}{0.9, 0.17, 0.31}
\definecolor{kellygreen}{rgb}{77, 186, 23}
\definecolor{azure}{rgb}{0.0, 0.5, 1.0}
\definecolor{gred}{rgb}{0.9, 0.17, 0.31}
\definecolor{gblue}{rgb}{0.0, 0.5, 1.0}
\definecolor{gyellow}{RGB}{244,180,0}
\definecolor{ggreen}{rgb}{0.3, 0.73, 0.09}
\definecolor{ggrey}{RGB}{115,115,115}

\DeclareMathOperator*{\argmin}{arg\,min}
\DeclareMathOperator*{\argmax}{arg\,max}
\newcommand{\error}[1]{\textcolor{gred}{\textbf{#1}}}

\newcommand{\reph}[1]{\textcolor{ggreen}{\textbf{#1}}}

\newcommand{\piref}{\pi_\text{ref}}

\title{Multi-Faceted Self-Consistent Preference Alignment for Query Rewriting in Conversational Search}

\author{Zhiyu Cao, Peifeng Li, Qiaoming Zhu\thanks{ \ \ Corresponding author} \\
        School of Computer Science and Technology, Soochow University, Suzhou, China  \\
        \texttt{zycao18@stu.suda.edu.cn}, \texttt{\{pfli, qmzhu\}@suda.edu.cn}
        }

\begin{document}
\maketitle
\begin{abstract}
Conversational Query Rewriting (CQR) aims to rewrite ambiguous queries to achieve more efficient conversational search. Early studies have predominantly focused on the rewriting in isolation, ignoring the feedback from query rewrite, passage retrieval and response generation in the rewriting process. To address this issue, we propose Multi-Faceted Self-Consistent Preference Aligned CQR (MSPA-CQR). Specifically, we first construct self-consistent preference alignment data from three dimensions (rewriting, retrieval, and response) to generate more diverse rewritten queries. Then we propose prefix guided multi-faceted direct preference optimization to learn preference information from three different dimensions. The experimental results show that our MSPA-CQR is effective in both in- and out-of-distribution scenarios.
\end{abstract}

\section{Introduction}

In Conversational Question Answering (CQA), it is frequently challenging to respond to user queries based exclusively on the conversational context. Conversational search emerges as an intermediate retrieval step, which requires the retrieval of relevant passages as external knowledge to assist in answering user queries. However, the user query may be ambiguous, resulting in the conversational search retrieving irrelevant passages. Consequently, existing approaches retrieve relevant passages by generating a decontextualized query via Conversational Query Rewriting (CQR) ~\cite{CONQRR,LLM4CS,AdaCQR,RETPO}, which can be combined with any well-established retriever for passage retrieval.

\begin{figure}[t]
\begin{center}
 \includegraphics[width=1\linewidth]{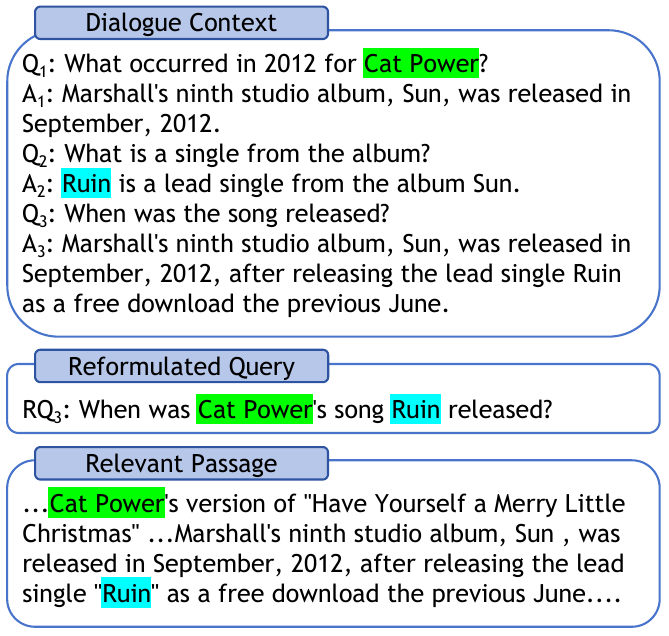}
 \vspace{-0.6cm}
 \caption{An example of CQR. Green and blue represent the key information for retrieval that was omitted in $\text{Q}_3$.}
 \label{fig:cqr_example}
\end{center}
\vspace{-0.3cm}
\end{figure}

As shown in Figure~\ref{fig:cqr_example}, the conversational search retrieves the relevant passage according to the query ``\textit{When was the song released?}''. However,  it is important to note that the meaning of this query cannot be understood correctly (e.g., what does ``\textit{song}'' refer to?) if the context of the conversation is left out. The corresponding rewritten query generated by CQR is RQ$_3$ in the figure, where the omitted keywords ``\textit{Cat Power}'' and ``\textit{Ruin}'' are crucial for retrieval.  The relevant passages that have been retrieved will assist the conversational system in generating the final response ``...Sun, was released in September, 2012...''.

Early studies on CQR ~\cite{T5QR,EDIRCS,ConvGQR} have utilized human-labeled rewritten queries as the ground truth for training. However, the models trained in this way are unable to  obtain feedback from passage retrieval and response generation, and they frequently serve to enhance the readability for humans, without directly benefiting retrieval processes.  Only a few studies have incorporated retrieval supervision signals during the rewriting process ~\citep{CONQRR,ConvGQR,RETPO}. However, they rely heavily on annotated gold passages to construct retrieval-relevant feedback information, which is time-consuming and labor-intensive.

 It should be noted that better-rewritten queries should correspond to better retrieval results and more reasonable responses. We expect the rewritten query in CQR to provide 
self-contained, critical retrieval-oriented and response-oriented textual information to achieve better retrieval results, which can benefit from different feedbacks, such as rewriting (rewritten query), retrieval (retrieved passages) and response (answer of the query). 
These feedbacks can provide information at different levels, thereby collaborating to promote the CQR task.

The objective of rewriting is to make the intent of the query as clear as possible, allowing the rewritten query to be \textit{self-contained without relying on the context of the conversation}. Therefore, feedback on rewriting can ensure that the rewritten query contains the complete user intent without ambiguous details. The feedback on retrieval tends to ensure that the rewritten query contains as much \textit{critical retrieval-oriented information} as possible, avoiding redundant information. This can guide the retrieval model to focus on the crucial content of the query, alleviating misleading effects caused by redundant information. The feedback on response can encourage the model to include \textit{response-oriented information} when generating the rewritten query. Since the retrieved passages usually contain key tokens that are helpful for response generation, this response-oriented information may appear in the final response, thus potentially affecting the final retrieval. Therefore, we are considering \textit{\textbf{how to use feedbacks from rewriting, retrieval, and response to facilitate CQR}}.

In this paper, we propose Multi-Faceted Self-Consistent Preference Aligned CQR (MSPA-CQR), consisting of two stages: multi-faceted preference data construction and prefix-guided multi-faceted preference optimization. In the first stage, we use LLMs to sample multiple rewritten queries in a few-shot manner. For each rewritten query, we retrieve relevant passages and generate corresponding response. To measure the quality of different rewrites, we designed an automated method of self-consistency scoring for the three preferences of rewriting, retrieval, and response to obtain their respective feedbacks. In the second stage, we perform Multi-Faceted Direct Preference Optimization (MDPO), aligning different dimensions by adding  preference types. Specifically, we add a prefix corresponding to each preference to the instructions during the MDPO process, so that the model can distinguish different preferences.

\begin{figure*}[t]
\begin{center}
 \includegraphics[width=1\linewidth]{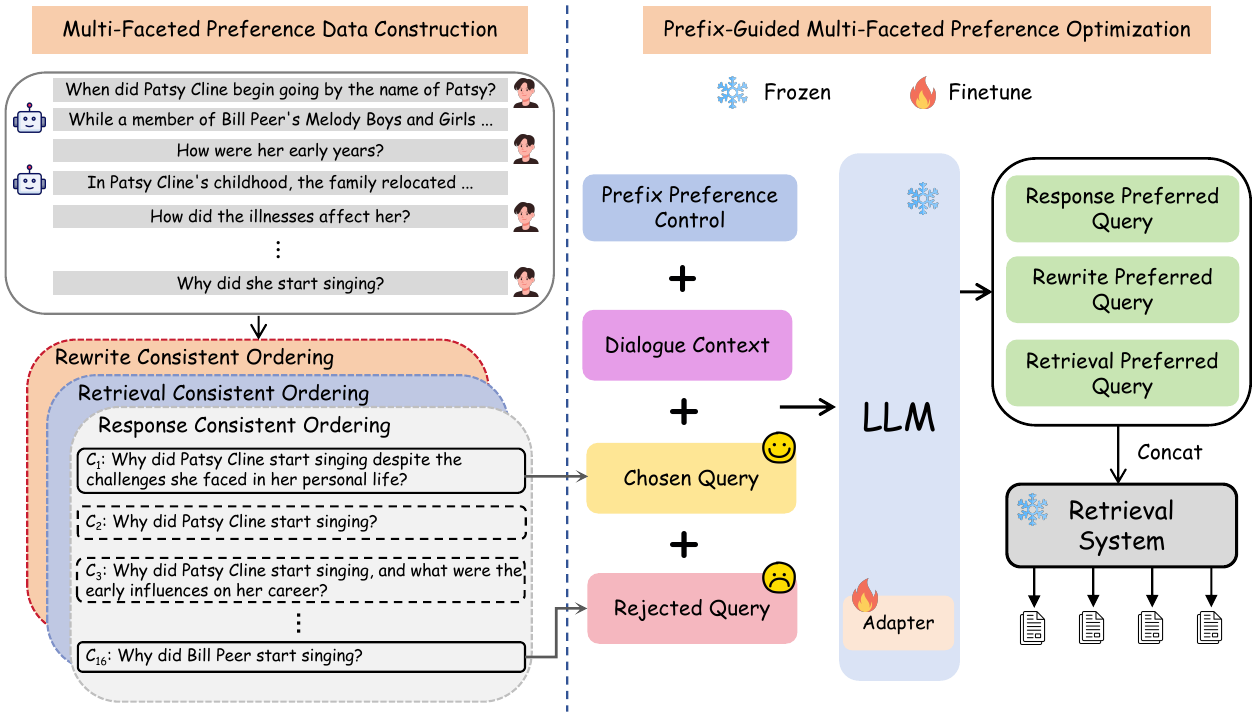}
 \caption{An overview of MSPA-CQR, including the two stages of Multi-Faceted Preference Data Construction and Prefix-Guided Multi-Faceted Preference Optimization.}
 \label{fig:method}
\end{center}
\end{figure*}

\section{Related Work}
Early studies on CQR relied on human-annotated rewritten queries to fine-tune the models ~\cite{T5QR,CONQRR,EDIRCS,ConvGQR}. Specifically, \citet{T5QR} utilized pre-trained language models to relax the independence assumption made when employing Maximum Likelihood Estimation (MLE) objectives in CQR. \citet{CONQRR} directly optimized rewritten queries to enhance retrieval performance. Since previous studies were inefficient in generating tokens, \citet{EDIRCS} selected tokens from dialogues and employed two search-oriented objectives. \citet{ConvGQR} designed a knowledge injection mechanism that optimized query rewriting through passage retrieval guidance and generated auxiliary potential answers to assist in retrieval.

However, human annotation is costly, and human-annotated rewritten queries often only enhance readability for humans, without necessarily benefiting retrieval. Recent studies have employed LLMs to generate rewritten queries \cite{LLM-Aided,LLM4CS,IterCQR,CHIQ,AdaCQR,RETPO}. Specifically, \citet{LLM-Aided} designed four important attributes to be included in the instructions, and used LLM as both query rewriters and rewrite editors. \citet{LLM4CS} proposed a prompting framework and introduced three prompting methods to use LLM for conversational search. IterCQR \cite{IterCQR} trained a CQR model through the direct utilization of information retrieval signals as rewards. Due to the ambiguous nature of conversation history, \citet{CHIQ} introduced LLMs to first enhance the conversation history, and then generate rewritten queries. \citet{AdaCQR} proposed aligning rewrite models from the perspectives of term and semantics. RETPO~\citep{RETPO} investigated the potential for rewriting through the utilization of diverse prompting techniques and performed preference-driven optimization based on feedback from the retriever.

Previous research only considered alignment from the retrieval preferences and did not take into account rewriting and response. Moreover, the construction of their preference data relied on human-annotated gold passages, which cannot be generalized to unlabeled data. In contrast, MSPA-CQR achieves self-consistent preference alignment from three dimensions, avoiding reliance on annotated gold passages.

\section{Methodology}

\subsection{Problem Formulation}
Given the current query $q_{t}$ and its dialogue context $\mathcal{H}_{t-1}=\left\{q_i, a_i\right\}_{i=1}^{t-1}$, where $q_i$ and $a_i$ represent the user's query and response in the $i$-th round, the goal of conversational search is to retrieve the top-$K$ most relevant passages $\mathcal{P}_t=\left\{p_j\right\}_{j=1}^K$ from a large passage collection $\mathcal{C}$ ($p_j \in \mathcal{C}$), which are used to generate the response $a_t$. Since the query $q_{t}$ may be ambiguous, CQR is often used as an intermediate step to rewrite $q_{t}$ into a rewritten query $\hat{q}_{t}$, making it self-contained, and then use the rewritten query $\hat{q}_{t}$ for passage retrieval. This process can be formally represented as follows:
\begin{equation}\label{eq:Formulation}
\begin{aligned}
\hat{q}_{t} &\leftarrow \mathcal{M}\left(\mathcal{H}_{t-1} \oplus q_{t}\right), \\
\mathcal{P}_t &\leftarrow \theta(\hat{q}_{t}, \mathcal{C}, K),
\end{aligned}
\end{equation}
where $\mathcal{M(\cdot)}$ represents the CQR model and the retriever $\theta(\cdot)$ can be a sparse or dense retriever (e.g., BM25 ~\cite{robertson2009probabilistic} or ANCE ~\cite{DBLP:conf/iclr/XiongXLTLBAO21}).

\subsection{Overview}
The proposed MSPA-CQR in Figure~\ref{fig:method} includes two stages: multi-faceted preference data construction and prefix-guided multi-faceted preference optimization. In the first stage, we construct preference data under the feedback of the three preferences of rewriting, retrieval, and response by scoring through self-consistency. Based on the results of consistency scoring, we then select the two samples with the highest and lowest scores as chosen and rejected samples, respectively. According to the multi-faceted preference data generated, in the second stage, we add preference prefixes before the MDPO instruction data to guide the model in learning rewritten queries under different preferences.

\subsection{Motivation}
\label{motivation}
The majority of previous studies exclusively focused on fine-tuning models through the utilization of human-annotated or model-generated rewritten queries, while only a few studies incorporated feedback from the retrieval preference. Nevertheless, feedback is imperative for CQR to retrieve more relevant passages and generate reasonable responses.

In this paper, we incorporate the feedback on three preferences of rewriting, retrieval, and response into the CQR task to combine their respective strengths.
Under the retrieval preference, rewritten query tends to retain only crucial information for retrieval while discarding redundant information. As illustrated in Figure~\ref{fig:method}, the query with the highest retrieval score is $C_2$. In $C_2$, not only entity coreference is resolved but also redundant information presented in $C_1$ is avoided, making it more favorable for retrieval. The feedback on rewriting tends to make the query self-contained, without relying on its context. In Figure~\ref{fig:method}, the query with the highest rewriting score is $C_3$, and it is a query that does not depend on the conversational context. 
The feedback on response prefers to include response-oriented  information in the rewritten query, such as ``\textit{despite the challenges she faced in her personal life}'' in $C_1$.

\subsection{Multi-Faceted Preference Data Construction}
\label{stage1}

We construct preference data around three dimensions: rewriting, retrieval, and response using LLM. Specifically, we first sample multiple candidate rewritten queries using LLM and then develop three different self-consistency scoring methods to score and rank the candidate rewritten queries according to the characteristics of the three preferences. Finally, chosen and rejected samples are obtained to construct multi-faceted preference dataset, which is prepared for the second stage of prefix-guided multi-faceted preference optimization.

For each query $q_t$ and its dialogue context $\mathcal{H}_{t-1}$, we first sample multiple candidate rewritten queries $RQ^{t} = \{rq_i^{t}\}_{i=1}^{K}$ from LLM (denoted as $f_{LLM}$) in a few-shot manner:
\begin{equation}
    rq_{i}^{t} \sim f_{LLM}(rw_{ins}^{i},\mathcal{H}_{t-1},q_t),
\end{equation}
where $rw_{ins}^{i}$ stands for the $i$-th rewrite instruction including five examples, which  can be found in Appendix~\ref{appendix:Prompt_Details}. To ensure diversity in sampling, we randomly select different five examples as demonstrations each time. 
For each rewritten query in $RQ^{t}$, we also use LLM to generate the response, thus collecting $RS^{t} = \{rs_i^{t}\}_{i=1}^{K}$ as follows:
\begin{equation}
    rs_{i}^{t} \sim f_{LLM}(rs_{ins},\mathcal{H}_{t-1},rq_{i}^{t}),
\end{equation}
where $rs_{ins}$ represents the response generation instruction, which can be found in Appendix~\ref{appendix:Prompt_Details}. In addition, for each rewritten query $rq_{i}^{t}$, we retrieve the top-100 most relevant passages, denoted as $P^{t} = \{\mathcal{P}_i^{t}\}_{i=1}^{K}$ ($|\mathcal{P}_i^{t}|=100$) as follows:
\begin{equation}
    \mathcal{P}_{i}^{t} \leftarrow \theta_{cand}(rq_{i}^{t}, \mathcal{C}, 100).
\end{equation}
Considering the retrieval efficiency, we use BM25 ~\cite{robertson2009probabilistic} as the retriever $\theta_{cand}$ here.

A better rewritten query should correspond to better retrieval results and a more reasonable response. Therefore, the next step is to score the elements in $RQ^{t}$ from the preferences of rewriting, retrieval, and response. We draw inspiration from the self-consistency strategy~\cite{self-consistency}, which sampled different reasoning paths and then used a majority vote as the final answer. One challenge in our scenario is that there may be multiple reasonable and different $rq_{i}^{t}$. Only considering frequency cannot guarantee semantic consistency. Therefore, we design different scoring methods based on three different preferences to obtain continuous scores for measuring self-consistency. The self-consistency score reflects feedback across different preferences.

In the context of rewriting and response, sentences that exhibit similar semantic content tend to demonstrate higher levels of consistency. With respect to retrieval, the larger the intersection of two sets of passages, the higher the consistency. Based on the above analysis, we define the following rewrite score $\mathbf{RW}$, retrieval score $\mathbf{RT}$, and response score $\mathbf{RP}$ from the perspective of self-consistency: 
\begin{equation*}
\small
\begin{aligned}
    \mathbf{RW}_{i}^{t} &= \frac{1}{K-1} \sum_{j \in [1, K], i \neq j} \textrm{NLI}(rq_{i}^{t}, rq_{j}^{t}) + \frac{\textrm{len}(rq_{i}^{t})}{\mathop{\textrm{max}}\limits_{j \in [1, K]} \textrm{len}(rq_{j}^{t})},\\
    &\mathbf{RT}_{i}^{t} = \frac{1}{K-1} \sum_{j \in [1, K], i \neq j} \left| \mathcal{P}_{i}^{t} \cap \mathcal{P}_{j}^{t} \right|, \\
    &\mathbf{RP}_{i}^{t} = \frac{1}{K-1} \sum_{j \in [1, K], i \neq j} \textrm{NLI}(rs_{i}^{t}, rs_{j}^{t}), \\
\end{aligned}
\end{equation*}
where NLI($\cdot$) denotes an NLI model that measures the semantic similarity. For two rewritten queries and their corresponding responses, a higher NLI-based semantic scoring indicates greater textual alignment between the two sentences. Regarding the passages retrieved by two rewritten queries, a larger intersection of retrieved passages suggests greater similarity in retrieval results. Notably, the rewrite score incorporates a penalty term defined as the ratio of query length to the maximum length among $K$ sampled queries. This penalty mechanism ensures that rewriting produces sufficiently comprehensive queries that function as standalone utterances containing adequate contextual information and prevents overly concise queries from achieving high rankings.

Based on the scores under the three preferences above, we obtain the corresponding chosen samples $rq_{+}^{t}$ and rejected samples $rq_{-}^{t}$ for all preferences as follows:

\begin{equation}
    rq_{+}^{t} = \argmax_{rq_{i}^{t} \in RQ^{t}}   \, \textrm{AP}_{i}^{t}, rq_{-}^{t} = \argmin_{rq_{i}^{t} \in RQ^{t}} \, \textrm{AP}_{i}^{t},
\end{equation}
where $\textrm{AP} \in \left\{\mathbf{RW}, \mathbf{RT}, \mathbf{RP}\right\}$. For example, in Figure~\ref{fig:method}, suppose 16 rewritten queries are sampled for the response preference, $C_1$ has the highest score, and $C_{16}$ has the lowest score. Hence, $C_1$ and $C_{16}$ form a pair of chosen and rejected samples for this preference. Finally, $rq_{+}^{t}$  and $rq_{-}^{t}$ contain three samples from the three preferences, respectively.

\begin{figure}[t]
\begin{center}
 \includegraphics[width=1\linewidth]{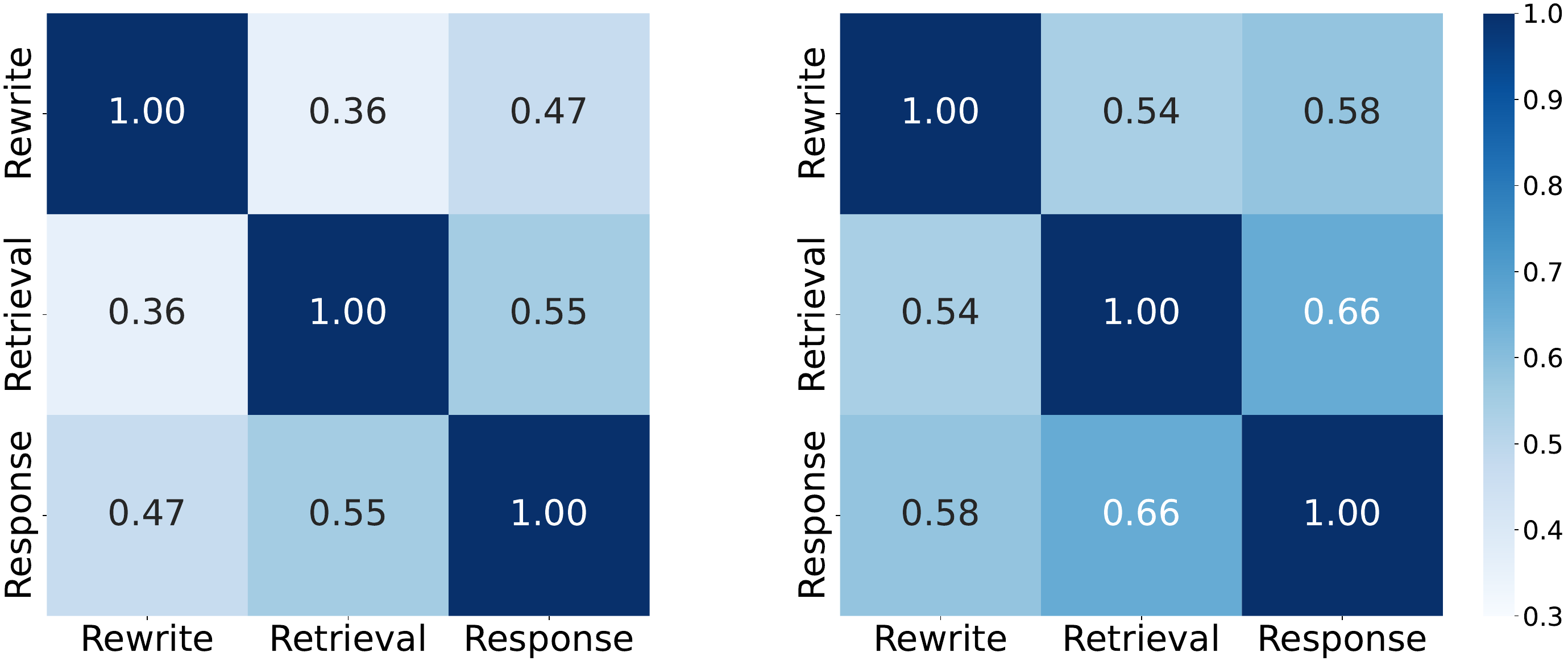}
 \caption{Kendall's Tau correlation between the consistency score ranking lists of different preferences on TopiOCQA (left) and QReCC (right).}
 \label{fig:kendall}
 \vspace{-0.2cm}
\end{center}
\end{figure}

\subsection{Prefix-Guided Multi-Faceted Preference Optimization}
We first conducted a preliminary experiment to observe the differences in preferences across three dimensions. To analyze the preferences of query across the three preferences of rewriting, retrieval, and response, we use Kendall-Tau ranking scores~\cite{kendall1938new} to measure the differences in the ranking of the scores across these three preferences. The results in Figure~\ref{fig:kendall} reveal significant differences between the three preferences. Specifically, the Kendall-Tau correlation between the rankings under rewriting and retrieval preferences on TopiOCQA is only 0.36, indicating that there are substantial differences in the query preferences.

To distinguish the three preferences, we define a set of preference tags $\mathcal{V}=$\{``[REWRITE]'', ``[RETRIEVAL]'', ``[RESPONSE]''\} according to three preferences, respectively.

Next, we collect a multi-faceted preference dataset $\mathcal D =\{(pr^i, x^i,rq_{+}^{i},rq_{-}^{i}) | pr^i \in \mathcal{V}\}_{i=1}^l$ containing $l$ instances where $rq_{+}^{i}$ is favored over $rq_{-}^{i}$ under the preference corresponding to $pr^i$. Here, $x$ represents the input sequence: $x = pr \,\oplus\, ins \,\oplus\, \mathcal{H} \,\oplus\, q$, where $ins$ denotes the query rewrite instruction (i.e., ``\texttt{Please rewrite the last query of the following conversation to make it more complete.}''), and $\oplus$ denotes the concatenation operation.

Given the multi-faceted preference dataset $\mathcal D$ collected above, the model needs to adaptively adjust between different preferences based on the prefix $pr$ in order to align with different preferences. Inspired by DPO~\cite{DPO}, we propose Multi-Faceted Direct Preference Optimization (MDPO) based on the dataset $\mathcal D$. We add a prefix corresponding to each preference to the instructions during the MDPO process, so that the model can distinguish between preferences from different dimensions.
Our optimization goal for MDPO with $\mathcal D$ is as follows:
\begin{equation}\label{eq:optimum_model2}
\small
\begin{aligned}
&\mathcal{L}_\text{MDPO}(\pi_{\theta}; \piref) = -\mathbb{E}_{(pr, x, rq_{+}, rq_{-})\sim \mathcal{D}} \\
&\left[\log \sigma \left(\beta \log \frac{\pi_{\theta}(rq_{+}\mid pr, x)}{\piref(rq_{+}\mid pr, x)}
- \beta \log \frac{\pi_{\theta}(rq_{-}\mid pr, x)}{\piref(rq_{-}\mid pr, x)}\right)\right],
\end{aligned}
\end{equation}
where $\hat{r}_\theta(pr, x, rq)=\beta \log \frac{\pi_\theta(rq \mid pr, x)}{\pi_{\mathrm{ref}}(rq \mid pr, x)}$ is the reward implicitly defined by the language model $\pi_\theta$ and the reference model $\pi_{\mathrm{ref}}$. Compared to vanilla DPO, our alignment process involves three different preferences and uses the prefix tag $pr$ to control preference optimization in different preferences. More insight on MDPO can be found in Appendix~\ref{appendix:MDPO}.

During the inference phase, for each test sample, we prepend preference tag to the instruction data, obtaining the in-context prompt. The model generates rewritten queries based on this prompt with corresponding preference guidance. Finally, we generate three queries with specific preferences, and then concatenate them for the final retrieval.

\begin{table*}[t!]
    \centering
    \small
    \begin{threeparttable}
    \begin{tabular*}{0.97\textwidth}{clccccccccc}
        \toprule
         \multirow{2}{*}{{\textbf{Type}}} & \multirow{2}{*}{{\textbf{System}}} & \multirow{2}{*}{{\textbf{Backbone}}} & \multicolumn{4}{c}{\textbf{TopiOCQA}} &  \multicolumn{4}{c}{\textbf{QReCC}}  \\
         \cmidrule(lr){4-7} \cmidrule(lr){8-11}
         & & & \textbf{MRR} & \textbf{NDCG} & \textbf{R@10}  & \textbf{R@100} & \textbf{MRR} & \textbf{NDCG} & \textbf{R@10}  & \textbf{R@100} \\
        \midrule 
        & \multicolumn{10}{g}{\textit{w/o Feedback}} \\
        \multirow{10}{*}{\rotatebox[origin=c]{90}{\textbf{Sparse (BM25)}}}
        & EDIRCS & T5-base & - & - & - & - & 41.2 & - & 62.7 & 90.2\\
        & LLM-Aided & ChatGPT & - & - & - & - & 49.4 & 46.5 & 67.1 & 88.2 \\
        & LLM4CS & ChatGPT & 27.9 & 26.4 & 48.4 & 71.1 & 51.6 & 49.3 & 75.3 & 92.6 \\
        & CHIQ & LLaMA2-7B & 25.6 & 23.5 & 44.7 & - & 54.3 & 51.9 & \textbf{78.5} & - \\
        & \multicolumn{10}{g}{\textit{w/ Retrieval Feedback}} \\
        & IterCQR & T5-base & 16.5 & 14.9 & 29.3 & 54.1 & 46.7 & 44.1 & 64.4 & 85.5 \\
        & AdaCQR & T5-base & 28.3 & 26.5 & 48.9 & 71.2 & 55.1 & 52.5 & 76.5 & 93.7 \\
        & RETPO & LLaMA2-7B  & 28.3 & 26.5 & 48.3 & 73.1 & 50.0 & 47.3 & 69.5 & 89.5 \\
        & \multicolumn{10}{g}{\textit{w/ Rewriting, Retrieval and Response Feedback}} \\
        & \cellcolor{mygray}{\textsc{MSPA-CQR}} & \cellcolor{mygray}LLaMA2-7B & \cellcolor{mygray}{\textbf{30.6}} & \cellcolor{mygray}{\textbf{29.5}} & \cellcolor{mygray}{\textbf{51.9}} & \cellcolor{mygray}{\textbf{75.2}} & \cellcolor{mygray}{\textbf{57.4}} & \cellcolor{mygray}{\textbf{54.1}} & \cellcolor{mygray}{77.6} & \cellcolor{mygray}{\textbf{95.2}} \\
        \midrule 
        & \multicolumn{10}{g}{\textit{w/o Feedback}} \\
        \multirow{10}{*}{\rotatebox[origin=c]{90}{\textbf{Dense (ANCE)}}}
        & EDIRCS & T5-base & - & - & - & - & 42.1 & - & 65.6 & 85.3\\
        & LLM-Aided & ChatGPT & - & - & - & - & 43.5 & 41.3 & 65.6 & 82.3 \\
        & LLM4CS & ChatGPT & 35.4 & 34.4 & 55.2 & 72.2 & 44.7 & 41.8 & 67.2 & 84.0 \\
        & CHIQ & LLaMA2-7B & 38.0 & 37.0 & 61.6 & - & 47.2 & 44.6 & 70.8 & - \\
        & \multicolumn{10}{g}{\textit{w/ Retrieval Feedback}} \\
        & IterCQR & T5-base & 26.3 & 25.1 & 42.6 & 62.0 & 42.9 & 40.2 & 65.5 & 84.1 \\
        & AdaCQR & T5-base & 38.5 & 37.6 & 58.4 & 75.0 & 45.8 & 42.9 & 67.3 & 83.8 \\
        & RETPO & LLaMA2-7B  & 30.0 & 28.9 & 49.6 & 68.7 & 44.0 & 41.1 & 66.7 & 84.6 \\
        & \multicolumn{10}{g}{\textit{w/ Rewriting, Retrieval and Response Feedback}} \\
        & \cellcolor{mygray}{\textsc{MSPA-CQR}} & \cellcolor{mygray}LLaMA2-7B & \cellcolor{mygray}{\textbf{41.4}} & \cellcolor{mygray}{\textbf{39.5}} & \cellcolor{mygray}{\textbf{63.5}} & \cellcolor{mygray}{\textbf{77.4}} & \cellcolor{mygray}{\textbf{48.7}} & \cellcolor{mygray}{\textbf{45.7}} & \cellcolor{mygray}{\textbf{72.3}} & \cellcolor{mygray}{\textbf{87.5}}
        \\
        \bottomrule 
    \end{tabular*}
    \end{threeparttable}
    \caption{
    Evaluation results of various retrieval system types on the test sets of TopiOCQA and QReCC, where all reported values of the baselines originated from their papers.
    }
    \label{table:main}
\end{table*}

\section{Experimentation}
\subsection{Experimental Setup}
\noindent \textbf{Datasets}
We conducted comprehensive experiments on two popular datasets  (Appendix~\ref{appendix:exp} for details): QReCC ~\cite{QReCC} and TopiOCQA ~\cite{TopiOCQA}.

\noindent \textbf{Retrieval Systems}
Following prior work in the CQR task ~\cite{IterCQR, RETPO}, we also evaluate our MSPA-CQR using sparse and dense retrieval systems. Sparse retrieval involves converting passages into sparse vectors for retrieval, while dense retrieval converts them into dense vectors. BM25 ~\cite{robertson2009probabilistic} and ANCE ~\cite{DBLP:conf/iclr/XiongXLTLBAO21} are used for sparse  and dense retrieval, respectively.

\noindent \textbf{Evaluation Metrics}
Following previous work ~\citep{AdaCQR,RETPO}, we used the following metrics  (Appendix~\ref{appendix:exp} for details): Mean Reciprocal Rank (MRR), Normalized Discounted Cumulative Gain (NDCG@3), and Recall@K (K = 10 or 100).

\noindent \textbf{Implementation Details}
The calculation of NLI scores uses \texttt{all-mpnet-base-v2}. Considering efficiency factors, the model used in the sampling module is \texttt{Qwen2.5-32B-Instruct-AWQ}. \texttt{Llama-2-7b-hf} is used as backbone and trained on LoRA~\cite{lora} fine-tuning with the rank of LoRA set to 8. Following previous work ~\cite{AdaCQR,RETPO}, we also adopted query expansion, where the prompt used is the same as that of response generation (as shown in Appendix~\ref{appendix:Prompt_Details}). For each query, a pseudo response generated by \texttt{Llama-2-7b-chat-hf} is appended to the query. More experimental details can be found in Appendix~\ref{appendix:exp}.

\noindent \textbf{Baselines}
The following strong models are used as baselines (Appendix~\ref{appendix:Baselines_Details} for details): EDIRCS~\citep{EDIRCS}, LLM-Aided~\citep{LLM-Aided}, LLM4CS~\citep{LLM4CS}, IterCQR~\citep{IterCQR}, CHIQ~\citep{CHIQ}, AdaCQR~\citep{AdaCQR}, RETPO~\citep{RETPO}.

\subsection{Main Results}
The experimental results are shown in Table~\ref{table:main}, our MSPA-CQR outperforms all baselines in both sparse and dense retrieval on almost all metrics, and the improvements are  significant with t-test at $p < 0.03$ over all compared baselines (except for R@10 on QReCC in the sparse setting). This results illustrate the effectiveness of our multi-faceted self-consistent preference alignment for the CQR task in conversational search.

Compared to the SOTA baselines (e.g., AdaCQR and RETPO) that used retrieval feedback for enhanced rewriting (\textit{w/ Retrieval Feedback}), our MSPA-CQR has achieved significant improvements. This is due to our method considering the rewriting, retrieval, and response feedback simultaneously using consistent scoring methods. 

For the methods without using feedback (\textit{w/o Feedback}), CHIQ achieved the best performance on most metrics. This is due to CHIQ using ranking list fusion, which integrates the ranking results of different modules through multiple retrievals. In contrast, our MSPA-CQR only requires one retrieval and outperforms CHIQ on all metrics except R@10 on QReCC in the sparse setting.

It is worth noting that IterCQR, AdaCQR, and RETPO used LLMs to generate rewritten queries, distilled into a relatively small model. Different from them, our MSPA-CQR obtains preference information from different dimensions rather than simply allowing the model to rigidly imitate the teacher LLM to generate a fixed rewritten query. MSPA-CQR can steer the model towards tailoring the generated rewritten queries from different perspectives, thus achieving better performance.

Compared to the baselines, MSPA-CQR has achieved significant improvements in MRR and NDCG. These two metrics focus on whether the retriever places more relevant passages at higher ranks.  The improvement in these two metrics indicates that the rewritten queries generated by our MSPA-CQR can effectively capture key information, thereby retrieving more relevant passages. This is attributed to the process of MDPO guided by prefix preference, where the model can distinguish useful information from irrelevant information from three preferences, generating more relevant rewritten queries.

We also evaluate the zero-shot generalization capability of MSPA-CQR and the results show that our MSPA-CQR outperforms all strong baselines on almost all metrics (Appendix~\ref{sec:zeroshot} for details).

\subsection{Analysis}
\label{sec:ablation}

\begin{table}[t]
    \centering
    \small
    \resizebox{\linewidth}{!}{
        \begin{tabular}{ccccccc}
        \hline Rewrite & Retrieval & Response & \textbf{MRR} & \textbf{NDCG} & \textbf{R@10}  & \textbf{R@100} \\
        \hline $\checkmark$ & $\checkmark$ & $\checkmark$ & 41.4 & 39.5 & 63.5 & 77.4 \\
        $\checkmark$ & $\checkmark$ & & 40.4 & 38.9 & 62.1 & 77.1 \\
        $\checkmark$ & & $\checkmark$ & 40.2 & 38.5 & 60.5 & 76.6 \\
        & $\checkmark$ & $\checkmark$ & 39.6 & 37.8 & 60.4 & 76.4 \\
        $\checkmark$ & & & 38.1 & 36.7 & 57.8 & 74.7 \\
        & $\checkmark$ & & 37.3 & 35.8 & 56.7 & 73.5 \\
        & & $\checkmark$ & 37.7 & 35.9 & 56.9 & 73.9 \\
        \hline
        \end{tabular}
    }
    \caption{
    Ablation study under the preference guidance of rewriting, retrieval, response, and the combinations.
    }
    \vspace{-2mm}
    \label{table:three_ablation}
\end{table}

\noindent \textbf{Preferences} During the inference phase, we combine rewritten queries generated under the guidance of the three preferences. To explore the effectiveness of each preference for retrieval, we conduct ablation experiments and the results are shown in Table~\ref{table:three_ablation}. The findings reveal that utilizing one or two preferences leads to a decline in performance. Furthermore, the efficacy of a rewritten query corresponding to a single preference is found to be less than that of rewritten queries under two preferences, suggesting that preferences of the three dimensions are complementary. 

The analysis further indicates that the use of a single type of preference, specifically the rewrite preference, yields optimal performance, suggesting its superior adaptability to retrieval processes. We also note that the retrieval performance of rewritten queries obtained from feedback under retrieval preference is the worst within these three categories, due to the fact that the passages contain some noisy information that does not match the intent of the current query, which in turn interferes with the rewriting process. In contrast, the rewrite and response preference focus on fine-grained information that is directly related to the current query, providing more precise feedback relevant to user intent.

\begin{table}[t]
    \small
    \centering
    \resizebox{\linewidth}{!}{
    \begin{tabular}{lcccc}
        \toprule
        & \multicolumn{4}{c}{\textbf{TopiOCQA}} \\
         \cmidrule(lr){2-5}
        \multicolumn{1}{c}{\textbf{Variant}} &  \textbf{MRR} & \textbf{NDCG} & \textbf{R@10}  & \textbf{R@100} \\        
        \midrule
        \cellcolor[gray]{0.9}MSPA-CQR (\textit{Ours}) & \cellcolor[gray]{0.9}41.4 & \cellcolor[gray]{0.9}39.5 & \cellcolor[gray]{0.9}63.5  & \cellcolor[gray]{0.9}77.4 \\
        \quad w/o. Prefix Guidance & 39.2 & 36.8 & 60.7 & 74.3 \\
        \quad w/o. MDPO & 29.1 & 27.8 & 48.6 & 65.8 \\
        \quad w/o. NLI & 40.8 & 38.7 & 62.7 & 76.8 \\
        \quad w/o. Query Expansion & 37.4 & 36.3 & 61.6 & 75.8 \\
        \midrule
        Aggregate-RRF  & 40.9 & 39.1 & 62.8 & 76.6 \\
        GPT-4o (Stage 1)  & 42.5 & 40.1 & 64.0 & 78.3 \\
        Qwen3-14B (Stage 2)  & 42.0	& 40.2	& 64.5	& 77.9 \\
        \bottomrule
    \end{tabular}
    }
    \caption{
    Ablation study for each component of \textsc{MSPA-CQR} and different LLMs.
    }
    \label{table:ablation}
    \vspace{-2mm}
\end{table}

\noindent \textbf{Components} To distinguish preferences along three different dimensions, we add prefix corresponding to each preference for training. To assess its impact, we remove these prefixes during the MDPO process (w/o. Prefix Guidance) and use vanilla DPO. During inference, three rewritten queries are obtained through three samplings for retrieval. As shown in Table~\ref{table:ablation}, removing preference prefixes leads to a significant decrease on all metrics. This also indicates the importance of distinguishing between the three preferences, which is more conducive to the model capturing the characteristics under different preferences. 

To further understand the impact of the MDPO process, we directly remove the entire MDPO process (w/o. MDPO). Similarly, we sample three times during inference and concatenate the sampled rewritten queries to eliminate the impact of sampling. As shown in Table~\ref{table:ablation}, there is a more significant decrease than ``w/o. Prefix Guidance'' in all metrics. This can also explain the importance of enabling the model to distinguish between relevant and irrelevant information in the MDPO process.

When calculating the consistency scores for rewriting and response, we used  NLI for scoring. Here we also tried using a majority voting method to select samples with the most and least votes as chosen and rejected samples (w/o. NLI). As shown in Table~\ref{table:ablation}, the majority voting is inferior to the NLI scoring, which may be attributed to the fact that the similarity between different rewritten queries and responses cannot be determined solely by surface-level consistency but requires deeper semantic information mining.

As shown in Table~\ref{table:ablation}, we tested the impact of query expansion on model performance (w/o. Query Expansion). Similar to observations from previous related studies on CQR~\cite{CHIQ,AdaCQR,RETPO}, query expansion can significantly improve retrieval performance. In addition, since our method generates queries with multiple different preference dimensions, it can also generate more diverse expanded answers.

\begin{figure}[t]
\begin{center}
 \includegraphics[width=1\linewidth]{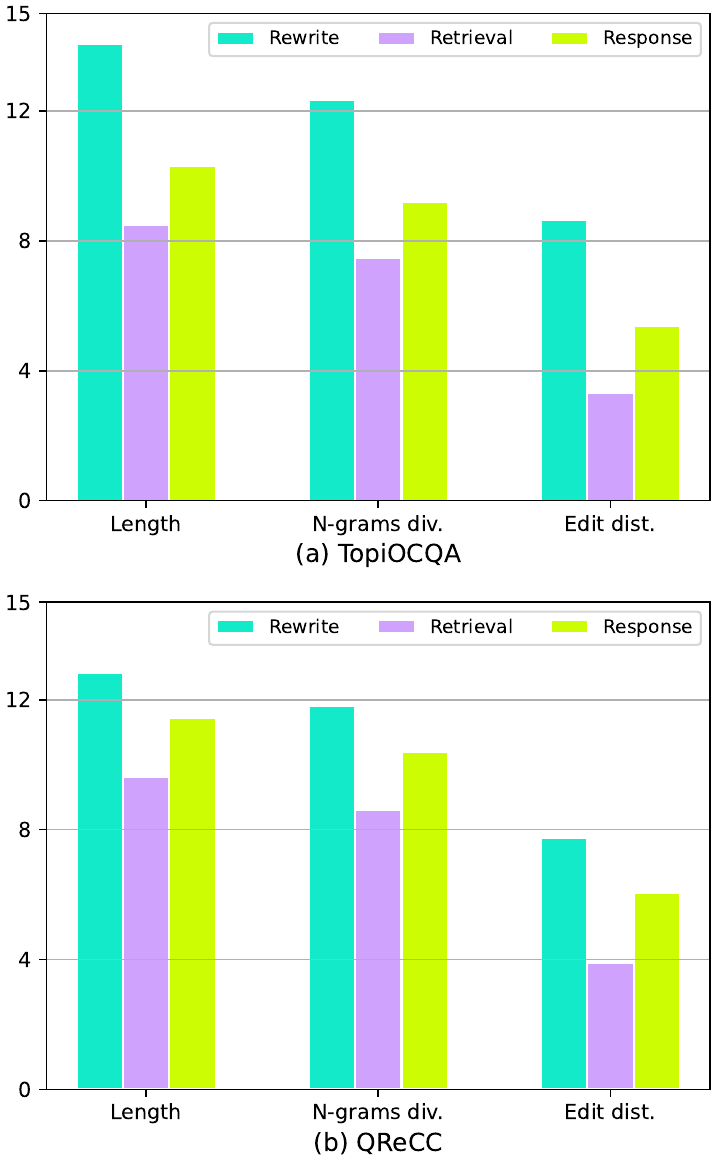}
 \caption{A comparison of the linguistic features of rewritten queries generated under three preference guides. ``N-grams div.'' refers to the diversity of n-grams ($\text{n}=2$), and ``Edit dist.'' refers to the Levenshtein distance.}
 \label{fig:Linguistic_features}
\end{center}
\vspace{-0.2cm}
\end{figure}

\noindent \textbf{Aggregate Strategy} 
We also tried the reciprocal rank fusion~\cite{RRF} (``Aggregate-RRF'' in Table~\ref{table:ablation}) to fuse queries in three dimensions (Appendix~\ref{sec:rrf} for details). Compared to direct concatenation in MSPA-CQR, the performance of RRF has decreased in all metrics. This may be due to the presence of noise in the query for some dimensions, and fusion of the retrieved results will amplify the noisy information. Whereas, direct concatenation can highlight the common information and reduce the effect of noisy information. Other fusion strategies can be found in Appendix~\ref{fa}.

\noindent \textbf{LLMs} To balance efficiency and performance, we used \texttt{Qwen2.5-32B-Instruct-AWQ} when sampling candidate rewritten queries. To analyze the impact of different models on sampling, we also tried using \texttt{GPT-4o-2024-08-06} for sampling. As shown in Table~\ref{table:ablation}, using GPT-4o for sampling can improve the performance because superior models are more conducive to obtaining high-quality preference data for the MDPO training. We also analyzed the effect of training a larger LLM \texttt{Qwen3-14B} to generate rewritten queries (Qwen3-14B (Stage 2)). As shown in Table~\ref{table:ablation}, the performance improved significantly, indicating that the performance scales with the size of the parameters.

\noindent \textbf{Further Analysis} Please refer to Appendix~\ref{fa}.

\subsection{Qualitative Analysis of Rewritten Queries under Three Preferences}
In order to achieve a more profound comprehension of the variances in rewritten queries guided by the three preferences, an analysis was conducted from three linguistic perspectives: length, n-gram diversity, and the edit distance between the original query and the rewritten query. The edit distance is measured using the Levenshtein distance \cite{levenshtein1966binary} . The experimental results are shown in Figure~\ref{fig:Linguistic_features}.

With regard to length, it is evident that the rewritten query under the rewrite preference is the longest, thereby ensuring the expression is more specific. Conversely, the query under the retrieval preference is comparatively shorter, a phenomenon that may be attributed to the necessity of retaining key information for retrieval while discarding superfluous information. Similarly, we also found that the rewrite and response preferences have higher n-gram diversity and greater edit distance from the original query. This finding indicates that these preferences introduce additional information, which may be more advantageous for the final retrieval.

\subsection{Intersection of Passages Retrieved under Different Preferences}
\begin{figure}[!t]
\begin{center}
 \includegraphics[width=1\linewidth]{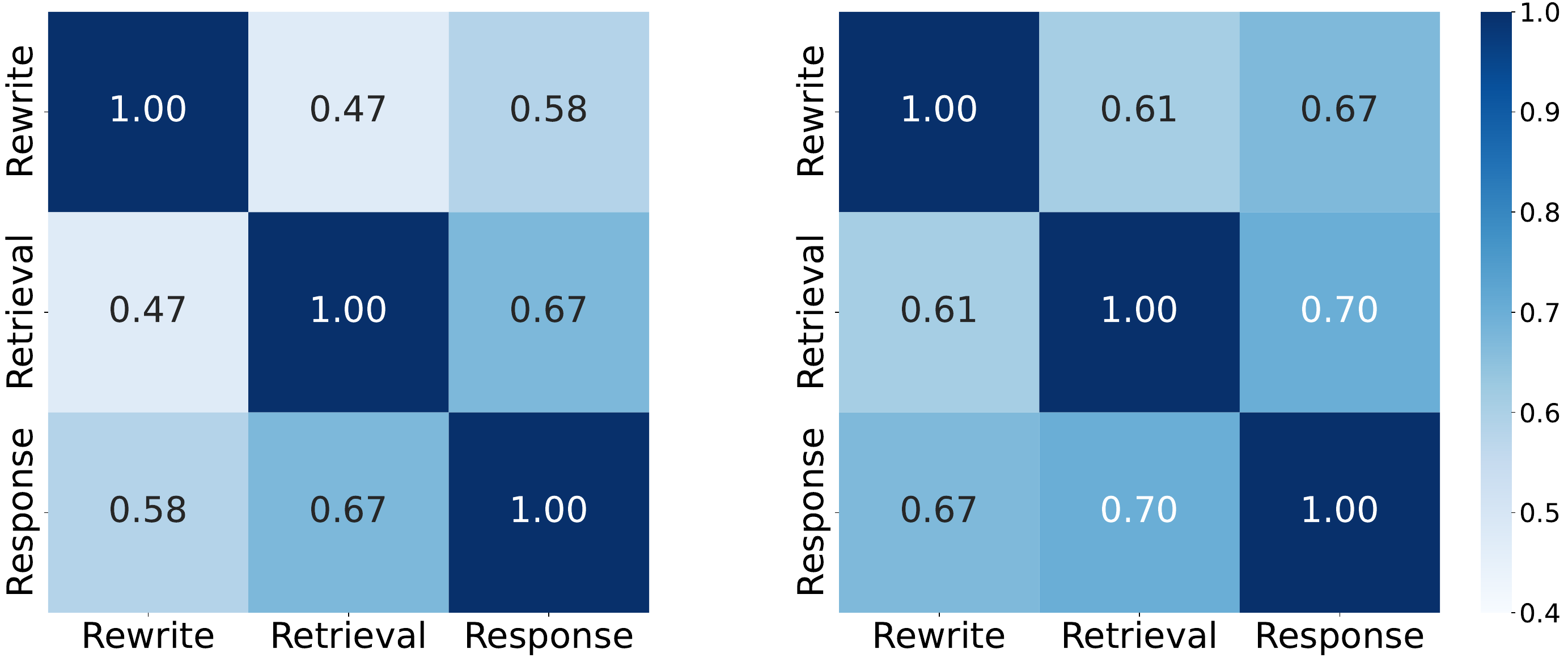}
 \caption{The intersection size of passages obtained through dense retrieval using rewritten queries guided by different preferences on TopiOCQA (left) and QReCC (right).}
 \label{fig:pair_intersection}
\end{center}
\vspace{-0.2cm}
\end{figure}

In ablation study, we found that the rewritten queries guided by different preferences have complementary effects. To further understand what makes them complementary, we analyzed the intersection size of passages retrieved under different preference guidance and the results are shown in Figure~\ref{fig:pair_intersection}. The results indicate that the intersection sizes are consistently below 70\%, suggesting the presence of significant information gaps between them. These findings imply that their combination  can potentially enhance the comprehensiveness of the retrieved information.

On the other hand, we also noticed that the intersection of the passages retrieved under the guidance of the rewrite and retrieval preferences is the smallest. In Section~\ref{sec:ablation}, we found that combining the rewritten queries generated under the rewrite and retrieval preferences yields the best performance among all pairwise combinations. This indicates that there is an information gap between different preferences, and the greater the information gap, the larger the complementary advantage.

\subsection{Case Study and Error Analysis}
To provide valuable insights for understanding the advantages of MSPA-CQR, we provide a case study in Table~\ref{table:case_study}. Green indicates information related to the original query's intent, while red indicates misleading information. 

The ground truth and response preferred query both interpret ``it'' as the ``Ned Kelly article'' that does not appear in the conversational context, and this caused interference to the retrieval. The query for retrieval preference retains the original query without introducing any additional auxiliary information. We noticed that only the response preferred query contains the key information ``Kelly took refuge''. However, due to the misleading information ``Ned Kelly article'', the key information in the query became sparse, resulting in no relevant passage being retrieved. By integrating the preferences from the three dimensions, the key information from the three preferences is integrated (i.e., ``shoemakers shop'', ``Kelly'', and ``took refuge''), and thus the gold passage is retrieved (ranked 19th among top-100). Please refer to Appendix~\ref{sec:examples} for more cases.

\begin{table}[t]
\small
\centering
\begin{tabularx}{\linewidth}{X}
\toprule
\textbf{Dialogue Context} \\
Q: What did Ned Kelly do? \\
A: An Australian bushranger, outlaw, gang leader and convicted police murderer. \\
Q: What happened after he was locked up for the night? \\
A: While he was escorted by four policemen, he absconded and ran, \reph{taking refuge} in a shoemaker's shop. \\
Q: Did it say what happened in the shoemakers shop? \\
\midrule
\textbf{Gold Passage}: On 18 September 1877 in Benalla, \reph{Kelly}, while drunk...he absconded and ran, \reph{taking refuge} in a \reph{shoemaker's shop}. The police and the shop owner tried to handcuff him but failed... \\
\midrule
\textbf{Ground Truth}: Did the \error{Ned Kelly article} say what happened to \reph{Kelly} in the \reph{shoemakers shop}? (\textbf{Rank: 79})\\
\midrule
\textbf{Rewrite Preferred Query}: Did it say what happened in the \reph{shoemaker's shop} when Ned \reph{Kelly} was there? (\textbf{Rank: Not Found})\\
\midrule
\textbf{Retrieval Preferred Query}: Did it say what happened in the \reph{shoemakers shop}? (\textbf{Rank: Not Found})\\
\midrule
\textbf{Response Preferred Query}: Did the \error{Ned Kelly article} say what happened in the \reph{shoemakers shop} after Ned \reph{Kelly} \reph{took refuge} there? (\textbf{Rank: Not Found})\\
\midrule
\textbf{Rewrite+Retrieval+Response}: \textbf{Rank: 19}\\
\bottomrule
\end{tabularx}
\caption{A case study on TopiOCQA, where ``Rank'' refers to the ranking of the retrieved gold passage, and ``Not Found'' means that there is no gold passage in the top-100 retrieved passages.}
\label{table:case_study}
\vspace{-0.2cm}
\end{table}

We conducted error analysis to understand the errors of MSPA-CQR. Taking TopiOCQA as an example, we found that among the samples where relevant passages were not retrieved from the top-100 passages by MSPA-CQR, 9.6\% of the samples could be retrieved by rewritten queries generated under one of these three individual preferences of rewriting, retrieval, or response. The samples that could be retrieved by the three individual preferences accounted for 3.6\%, 3.2\%, and 4.1\% respectively. We found that this was due to interference caused by errors in the rewritten queries guided by one or two of the preferences, leading to the failure to retrieve relevant passages. How to mitigate the interference caused by some preferences is also a direction worth exploring in the future.

\section{Conclusion}
In this paper, we propose a novel framework, Multi-Faceted Self-Consistent Preference Aligned CQR~(MSPA-CQR), which operates along three dimensions: rewriting, retrieval, and response. The framework consists of two stages: multi-faceted preference data construction and prefix-guided multi-faceted preference optimization. Experimental results on five datasets demonstrate that MSPA-CQR outperforms state-of-the-art baselines in both in- and out-of-distribution scenarios.

\section*{Limitations}
Although our proposed MSPA-CQR can promote more effective conversational search, it also has the following three limitations. Firstly, there is still room for optimization in the consistency scoring of the preference data, and it is worth considering the design of some heuristic scoring methods in the future. Secondly, in the preference optimization stage, we directly mix the data of the three preferences for training. However, there may be certain correlations between the three preferences, and we can design some methods to promote them mutually. Finally, we simply use DPO for preference optimization without trying other effective alignment methods.

\section*{Acknowledgements}
The authors would like to thank the anonymous reviewers for their comments on this paper. This research was supported by the National Natural Science Foundation of China (Nos. 62276177 and 62376181), and Project Funded by the Priority Academic Program Development of Jiangsu Higher Education Institutions.

\bibliography{custom}

\newpage
\appendix

\section{Prompt Details}
\label{appendix:Prompt_Details}
The prompt used in sampling candidate rewritten queries is as follows. ``Annotated Sample'' is the sample that needs to be annotated with LLM, i.e. ``Dialogue'' and ``Rewritten Sentence'' below. To ensure diversity in the rewritten queries obtained through sampling, we randomly select 5 samples as examples each time.
\begin{tcolorbox}[colback=white,colframe=blue!95!black,title=Prompt used in sampling candidate rewritten queries]
Please rewrite the last statement of the following dialogue to make it more complete. Just provide the rewritten sentence without any additional content.\\
Demonstrations: \\
\{\texttt{Five examples}\} \\
Annotated Sample: \\
Dialogue: \\
\{\texttt{Dialogue context}\} \\
Rewritten Sentence:
\end{tcolorbox}

The prompt used in response generation is as follows. For each rewritten query, we use a fixed set of 5 examples as demonstrations to add to the prompt. This is to ensure that the prompt for generating a response to each query is consistent, avoiding the impact of different prompts.
\begin{tcolorbox}[colback=white,colframe=blue!95!black,title=Prompt used in response generation]
Given a question, please answer the question in a sentence. The answer should be as informative as possible. \\
Demonstrations:
\{\texttt{Five examples}\} \\
Annotated Sample: \\
Question:
\{\texttt{Candidate rewritten query}\} \\
Answer:
\end{tcolorbox}

\section{Experimental Settings}
\label{appendix:exp}
\paragraph{Details of Datasets} The specific statistical information of the datasets is shown in Table~\ref{table:datasets}. Compared to QReCC, the conversations in TopiOCQA involve many cases of topic shifts. CAsT-19 and CAsT-20 share the same collection with 38M passages. The collection for CAsT-21 contains more passages (40M). Since we generate three pairs of rewritten queries with different dimensions for each training sample, the amount of preference alignment data is three times the original training data volume given in Table~\ref{table:datasets}. For the validation set, following previous work~\cite{AdaCQR}, we extracted 800 samples from the training set as the validation set.

\paragraph{Details of Metrics} MRR calculates the average reciprocal rank of the first relevant passage for each query. NDCG@3 evaluates the top-3 retrieval results by considering relevance and ranking. Recall@K is used to assess the system's ability to retrieve relevant passages within the top-K retrieval results. We use the \texttt{pytrec\_eval} tool to calculate these metrics.

\paragraph{Implementation Details} MSPA-CQR performs one epoch of supervised fine-tuning on the original dataset as a warm-up, and then trains for three epochs on the preference dataset we constructed. During the supervised fine-tuning and preference optimization stages, the learning rate is set to 2e-5, with batch size and gradient accumulation steps set to 8 and 2 respectively. We use Faiss~\citep{Faiss} and Pyserini~\citep{Pyserini} for dense retrieval and sparse retrieval respectively. For BM25, $k_1$ and $b$ are set to 0.9 and 0.4 on TopiOCQA, and 0.82 and 0.68 on QReCC, where $k_1$ controls the non-linear term frequency normalization, and $b$ is the scale of the inverse document frequency.

\begin{table}[t]
\centering
\small
\setlength{\tabcolsep}{4pt}{
\begin{tabular}{llrrr}
\toprule
Dataset & Split & \#Conv. & \#Turns(Qry.) & \#Collection \\ \midrule
\multirow{2}{*}{TopiOCQA} & Train & 3,509 & 45,450 & \multirow{2}{*}{25M} \\
 & Test  & 205 & 2,514 & \\
\midrule
\multirow{2}{*}{QReCC} & Train & 10,823 & 63,501 & \multirow{2}{*}{54M} \\
 & Test  & 2,775 & 16,451 & \\
\midrule
{CAsT-19} & Test  & 50 & 479 & \multirow{2}{*}{{38M}} \\
{CAsT-20} & Test  & 25 & 208 & \\ 
\midrule
{CAsT-21} & Test  & 26 & 239 & 40M \\ 
\bottomrule
\end{tabular}}
\caption{Statistics of conversational search datasets.}
\vspace{-2ex}
\label{table:datasets}
\end{table}

\begin{table*}[!t]
\centering
\resizebox{\linewidth}{!}{
\begin{tabular}{lcccccccccc}
\toprule
\multicolumn{1}{l}{\multirow{2}{*}{System}} & \multirow{2}{*}{Backbone} & \multicolumn{3}{c}{CAsT-19} & \multicolumn{3}{c}{CAsT-20} & \multicolumn{3}{c}{CAsT-21} \\ 
\cmidrule(lr){3-5} \cmidrule(lr){6-8} \cmidrule(lr){9-11} 
& & MRR & N@3  & R@10  & MRR & N@3 & R@10 & MRR     & N@3 & R@10  \\ 
\midrule
E5-Mistral & Mistral-7B & 62.2 & 31.3 & 9.5 & 22.0 & 15.4 & 8.4 & 48.2 & 32.5 & 20.5\\
HyDE & ChatGPT-3.5 & 55.6 & 39.2 & 10.0 & 44.8 & 29.3 & 16.9 & - & - & -\\
Query2doc & ChatGPT-3.5 & 58.8 & 42.4 & 11.6 & 48.6 & 32.5 & 17.3 & - & - & -\\
InstructorR & ChatGPT-3.5 & 61.2 & 46.6 & 10.4 & 43.7 & 29.6 & 8.3 & 46.7 & 32.5 & 18.4 \\
LLM4CS & ChatGPT-3.5 & 70.4 & 46.8 & 11.7 & \textbf{58.6} & 41.5 & 19.3 & 66.1 & 46.9 & 24.4\\
RepLLaMA & LLaMA2-7B & 62.4 & 31.6 & 10.6 & 26.8 & 18.3 & 10.4 & 47.4 & 32.7 & 19.6 \\
LLM-Embedder & LLaMA2-7B & 63.3 & 36.6 & 11.4 & 25.2 & 15.4 & 8.7 & 46.8 & 31.2 & 17.3\\
\fname{} &  LLaMA2-7B & 73.3 & 50.5 & 12.9 & 54.0 & 38.0 & 19.3 & 62.9 & 46.5 & 25.2 \\
AdaCQR &  T5-base & 74.5 & - & 13.8 & 56.6 & - & 19.2 & 64.2 & - & 25.0 \\
\midrule
MSPA-CQR &  LLaMA2-7B & \textbf{76.1} & \textbf{52.8} & \textbf{14.6} & 58.5 & \textbf{43.5} & \textbf{19.9} & \textbf{67.4} & \textbf{47.1} & \textbf{26.4} \\
        \bottomrule
     \end{tabular}}
     \caption{Zero-shot retrieval performances under the dense retrieval (ANCE).}
     \label{tab:cast_main}
\end{table*}

\section{Insights of MDPO}
\label{appendix:MDPO}
From an implementation perspective, MDPO indeed distinguishes preference data across different dimensions by adding preference prefixes based on the DPO approach. However, we have improved MDPO over DPO from both training and inference perspectives. From the training perspective, our designed preference prefixes enable the model to differentiate rewritten queries corresponding to different dimensions. From the inference perspective, by defining different preference dimension prefixes, the model can be guided to generate rewritten queries under various preferences. In contrast, DPO only generates responses corresponding to a single preference.

\section{Details of Baselines}
\label{appendix:Baselines_Details}
The details of the baselines used in this paper are as follows.

(1) \textbf{EDIRCS}~\cite{EDIRCS} simultaneously selected tokens from the conversation history and generated new tokens, using two search-oriented objectives to enhance learning.

(2) \textbf{LLM-Aided}~\cite{LLM-Aided}  used LLMs as query rewriter and rewrote editors by providing explicit instructions with four desirable properties.

(3) \textbf{LLM4CS}~\cite{LLM4CS} explored three prompting strategies to generate multiple query rewrites and hypothetical responses, which were then aggregated.

(4) \textbf{IterCQR}~\cite{IterCQR} used an iterative framework to alternate between generating candidate queries and optimizing the CQR model, with information retrieval signals as rewards.

(5) \textbf{RETPO}~\cite{RETPO} reformulated search queries based on the preferences of the target retrieval system.

(6) \textbf{AdaCQR}~\cite{AdaCQR} used contrastive loss from the perspective of terms and semantics.

(7) \textbf{CHIQ}~\cite{CHIQ} proposed a two-step method for query rewriting based on open-source language models: enhancing dialogue history and then generating search queries.

(8) \textbf{RepLLaMA}~\cite{RepLLaMA} fine-tuned the LLaMA model, serving as both a dense retriever and a pointwise re-ranker.

(9) \textbf{E5-Mistral}~\cite{E5-Mistral} generated diverse synthetic data for tens of thousands of text embedding tasks in 93 languages.

(10) \textbf{LLM-Embedder}~\cite{LLM-Embedder} integrated four key retrieval capabilities: knowledge, memory, examples, and tools.

(11) \textbf{HyDE}~\cite{HyDE} decomposed dense retrieval into two tasks: a generation task performed by a language model following instructions, and a document-document similarity task performed by a contrastive encoder.

(12) \textbf{Query2doc}~\cite{Query2doc} generated pseudo-documents by few-shot prompting LLMs, and connected them with the original query to form a new query.

(13) \textbf{InstructorR}~\cite{InstructGPT} designed three instructing strategies to calculate the session-passage relevance score.

\section{Implementation of Reciprocal Rank Fusion}
\label{sec:rrf}
Given the passage collection $\mathcal{C}$ and the retrieval results $R$ corresponding to three dimensions, Reciprocal Rank Fusion (RRF) ranks documents based on the following score,
\begin{equation}
RRFscore(d \in \mathcal{C})=\sum_{r \in R} \frac{1}{k+r(d)},
\end{equation}
where $k$ is a constant that we set to 60, and $R$ represents the retrieval results of three preference dimensions.

\section{Zero-Shot Analysis}
\label{sec:zeroshot}
To test the zero-shot generalization ability of our proposed method, we also conducted experiments on three other datasets TREC CAsT 19-21 ~\citep{CAsT-2019,CAsT-2020,CAsT-2021}.  For the evaluation of zero-shot generalization, we additionally compared it with the following methods: RepLLaMA \cite{RepLLaMA}, E5-Mistral \cite{E5-Mistral}, LLM-Embedder \cite{LLM-Embedder}, HyDE \cite{HyDE}, Query2doc \cite{Query2doc} and InstructorR \cite{InstructorR}, as mentioned in Appendix~\ref{appendix:Baselines_Details}.
 
 The results are shown in Table~\ref{tab:cast_main}. We directly transfer our model trained on QReCC to these three datasets. Compared to the baselines, our MSPA-CQR achieves better generalization. 
It is worth noting that AdaCQR uses T5-base as the backbone, yet surpasses the baselines on ChatGPT-3.5 and LLaMA2-7B. This may be because larger models are more prone to overfitting the training set, making it difficult to generalize in zero-shot scenarios. However, our MSPA-CQR not only performs well in in-distribution cases, but its performance on CAsT 19-21 also demonstrates that MSPA-CQR can generalize to out-of-distribution scenarios.

\begin{table}[t]
    \small
    \centering
    \resizebox{\linewidth}{!}{
    \begin{tabular}{lcccc}
        \toprule
        & \multicolumn{4}{c}{\textbf{TopiOCQA}} \\
         \cmidrule(lr){2-5}
        \multicolumn{1}{c}{\textbf{Variant}} &  \textbf{MRR} & \textbf{NDCG} & \textbf{R@10}  & \textbf{R@100} \\        
        \midrule
        \cellcolor[gray]{0.9}MSPA-CQR (\textit{Ours}) & \cellcolor[gray]{0.9}41.4 & \cellcolor[gray]{0.9}39.5 & \cellcolor[gray]{0.9}63.5  & \cellcolor[gray]{0.9}77.4 \\
        \quad w/o. Length Penalty Term & 41.1	& 39.0	& 62.8	& 76.9 \\
        \quad w/ passages & 41.8	& 40.2	& 64.1	& 77.6 \\
        \midrule
        MSPA-CQR (CombGMNZ)	& 40.5	& 38.8	& 62.6	& 76.8 \\
        MSPA-CQR (Log\_ISR)	& 40.7	& 39.1	& 62.4	& 76.5 \\
        MSPA-CQR (Marginal Probability of Answer)	& 41.1	& 39.2	& 63.8	& 77.6 \\
        MSPA-CQR (ColBERT)	& 43.6	& 42.1	& 66.2	& 79.8 \\
        Llama-2-7b-chat-hf (Stage 1)	& 39.7	& 37.6	& 61.2	& 75.3 \\

        \bottomrule
    \end{tabular}
    }
    \caption{
    Ablation experiments of different modules in the MSPA-CQR.
    }
    \label{table:appendix:ablation}
    \vspace{-2mm}
\end{table}

\section{Further Analysis}
\label{fa}
\subsection{Analysis on Length Penalty Term} 
Taking dense retrieval on TopiOCQA as an example, we tried to remove this length penalty term. The experimental results are shown in Table~\ref{table:appendix:ablation}, the performance on all the metrics decreases after removing this length penalty term, which indicates that the length penalty term can promote the model to generate more complete rewritten queries, which in turn improves the performance of retrieval.

\subsection{Other Query Fusion Strategies} 
Due to the additional computational overhead introduced by learnable fusion strategies, we used a training-free retrieval fusion method. We also experimented with using other retrieval fusion methods, and the experimental results of dense retrieval on TopiOCQA using CombGMNZ~\cite{CombGMNZ} and Log\_ISR~\cite{Log_ISR} as examples are shown in Table~\ref{table:appendix:ablation}. Using CombGMNZ and Log\_ISR both lead to a decline in retrieval performance, which may be due to the fact that both of these retrieval result level fusion methods introduce noisy information retrieved by certain queries, which interferes with the final retrieval results. Future research can focus on optimizing model preferences while allowing the model to assess the reliability of each dimension and integrate retrieval results based on the assessment.

\subsection{Preference Data Generation Using Other Models} In addition to using \texttt{Qwen2.5-32B-Instruct-AWQ} and \texttt{GPT-4o} to generate candidate rewritten queries, we also experimented with using \texttt{Llama-2-7b-chat-hf} for generation, as shown in Table~\ref{table:appendix:ablation}. It can be observed that there was a significant decline in performance (a decrease of 2.1 on R@100), indicating that data generated by stronger models has higher quality and is more conducive to aligning the model's preferences.

\subsection{Adaptation to Other Retrievers} Since we use models and retrievers for quantitative evaluation, some biases are inevitably introduced. However, it is noteworthy that we use Qwen2.5-32B-Instruct-AWQ in the data construction stage, while the LLM chosen for the training and inference stages is LLaMA2-7B. Although they belong to different model families, MSPA-CQR can still achieve excellent performance. Additionally, we use BM25 as the retriever in the data construction stage, and using ANCE, a dense model, for retrieval in the inference stage still demonstrates good generalization.

To further verify the generalization of MSPA-CQR, we also used ColBERT as a retriever for evaluation. As shown in Table~\ref{table:appendix:ablation}, the performance of MSPA-CQR achieved further improvement after using ColBERT for retrieval (i.e., ``MSPA-CQR (ColBERT)''). Therefore, MSPA-CQR can adapt to different retrieval systems.

\subsection{Different Retrieval Feedback Methods}
To verify the effectiveness of our designed method using self-consistency scores to reflect retrieval feedback, we also experimented using AdaQR~\cite{AdaQR}'s reward (i.e., marginal probability of answer) as our feedback score for retrieval dimensions. As shown in Table~\ref{table:appendix:ablation}, the retrieval feedback method we proposed is almost on par with the method in AdaQR. However, the method in AdaQR relies on annotated answers in the dataset, while our self-consistency scoring approach does not depend on manual annotations.

\begin{figure}[t]
    \centering
   \begin{tcolorbox}[
      colback=white, colframe=black, arc=3mm, width=\linewidth,
      title=\textbf{Response Preference Data Construction}, 
      coltitle=white, colbacktitle=gray, fonttitle=\bfseries
    ]
    Given the following conversation history, the current query, and three passages related to the current query, please generate a response for the current query. You only need to output the response, please do not output any extra content.
    
    Conversation History: {Conversation History}
    
    Current Query: {Current Query}
    
    Relevant Passages: 
    
    \textbf{Passage 1}: 
    
    \{Relevant Passage 1\}
    
    \textbf{Passage 2}: 
    
    \{Relevant Passage 2\}
    
    \textbf{Passage 3}: 
    
    \{Relevant Passage 3\}
    
    Response:
    \end{tcolorbox}
    \caption{The prompt used to construct the response preference data based on retrieval.}
    \label{fig:retrieval_response_pref}
\end{figure}

\subsection{Retrieval-based Construction of Response Preference Data}
Since response generation based on retrieved passages would greatly increase the computational resource consumption for data construction, we adopted the approach of generating responses directly based on rewritten queries. We also evaluated the strategies of response generation and preference data construction based on retrieved passages. Specifically, we retrieve the three most relevant passages based on the rewritten query and add these three passages to the prompt for the model to generate responses. The prompt is shown in Figure~\ref{fig:retrieval_response_pref}. As shown in the Table~\ref{table:appendix:ablation}, the model performance is improved after adding the retrieved passages (i.e., ``w/ passages''). This is due to the fact that the responses generated by the model based on the retrieved passages are closer to the semantics of the query, which better reflects the response preferences and thus constructs higher quality preference data. However, the addition of retrieved passages consumes more computational resources for data construction, and it is worth exploring how to balance the efficiency and effectiveness of data construction in the future.

\begin{table}[t]
    \small
    \centering
    \resizebox{\linewidth}{!}{
    \begin{tabular}{lccc}
        \toprule
        & \multicolumn{1}{c}{Response Generation} & \multicolumn{1}{c}{Sparse Retrieval}  & \multicolumn{1}{c}{Dense Retrieval} \\
        \midrule
        TopiOCQA  & 0.74 & 0.82 & 0.77 \\
        QReCC  & 0.79	& 0.84	& 0.81 \\
        \bottomrule
    \end{tabular}
    }
    \caption{
    Kendall's Tau correlation between self-consistency scores and downstream task performance.
    }
    \label{table:corr_dt}
    \vspace{-3mm}
\end{table}

\subsection{Correlation between Self-Consistency Scores and Downstream Tasks} 
We measure preferences in different dimensions through self-consistency scores, based on the assumption that self-consistency is closely related to performance on downstream tasks. To verify this hypothesis, we measured the correlation between the self-consistency score and downstream task performance. For response generation, we calculate the BLEU$_1$, BLEU$_2$, ROUGE$_1$, and ROUGE$_2$ scores between the generated response and ground truth, and sum them up to measure the performance of response generation. For sparse retrieval and dense retrieval, we respectively sum up MRR, NDCG@3, R@10, and R@100 to measure the performance of retrieval. As shown in Table~\ref{table:corr_dt}, the self-consistency scores is closely related to the performance of downstream tasks (all greater than 0.7), indicating that our designed self-consistency scoring method adequately represents preferences across various dimensions.

\begin{figure}[t]
\begin{center}
 \includegraphics[scale=0.5]{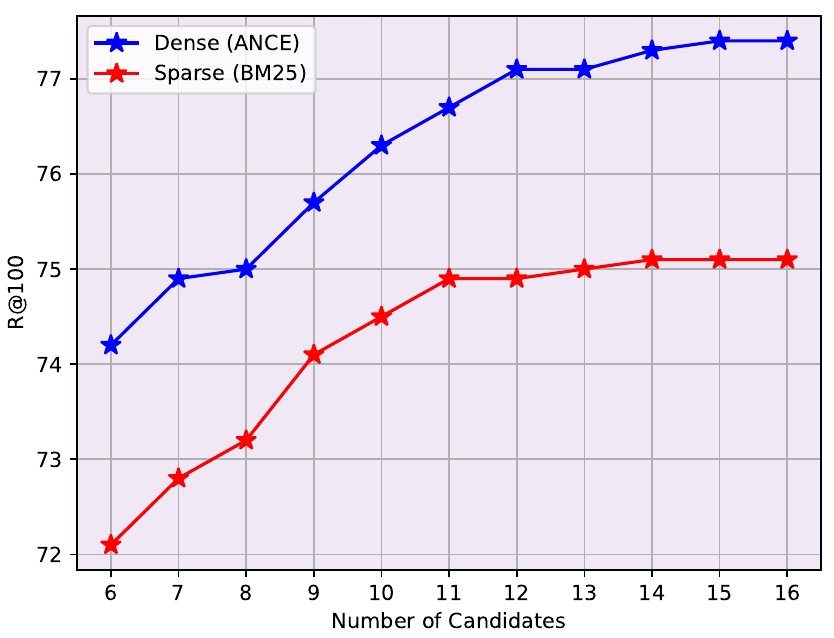}
 \caption{The trend of R@100 on TopiOCQA with the change in the number of sampled candidate rewritten queries.}
 \label{fig:num_cand}
\end{center}
\vspace{-0.4cm}
\end{figure}
\begin{figure}[t]
\begin{center}
 \includegraphics[scale=0.5]{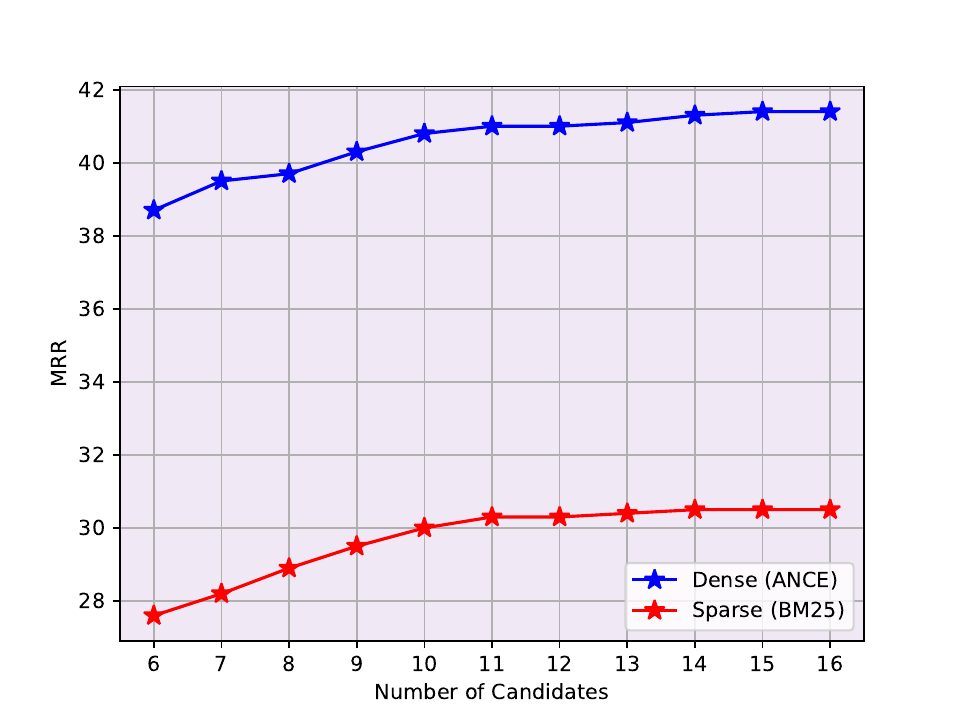}
 \caption{The trend of MRR on TopiOCQA with the change in the number of sampled candidate rewritten queries.}
 \label{fig:num_cand_MRR}
\end{center}
\vspace{-0.4cm}
\end{figure}

\subsection{Number of Sampled Rewritten Queries}
As shown in Figure~\ref{fig:num_cand}, we explore the impact of the number of sampled candidate rewritten queries in the first phase on R@100. It can be observed that in both sparse and dense retrieval scenarios, R@100 exhibits an upward trend in conjunction with an increase in the number of samples. This phenomenon can be attributed to the enhanced precision in the calculation of consistency scores that results from a larger sample size. Consequently, this facilitates the selection of higher-quality chosen and rejected samples for the MDPO training. Additionally, when the number of sampled rewritten queries reaches a certain threshold, performance gradually saturates. At this point, the consistency scores can accurately reflect the sample quality under each preference.

In addition to the trend on R@100, Figure~\ref{fig:num_cand_MRR} shows the MRR trend with the number of sampled rewritten queries. It can be observed that MRR also exhibits an upward trend  as the number of sampled rewritten queries increases. For example, in dense retrieval, as the sample size increased from 6 to 16, MRR is improved by 2.7.

\begin{table}[t]
    \small
    \centering
    \resizebox{\linewidth}{!}{
    \begin{tabular}{lcccc}
        \toprule
        & \multicolumn{4}{c}{\textbf{TopiOCQA}} \\
         \cmidrule(lr){2-5}
        \multicolumn{1}{c}{\textbf{Different Values of T}} &  \textbf{MRR} & \textbf{NDCG} & \textbf{R@10}  & \textbf{R@100} \\        
        \midrule
        MSPA-CQR (T = 10)	& 40.7	& 38.4	& 62.7	& 76.8 \\
        MSPA-CQR (T = 50)	& 41.6	& 39.8	& 64.1	& 78.2 \\
        MSPA-CQR (T = 100)	& 41.4 & 39.5 & 63.5  & 77.4 \\
        \bottomrule
    \end{tabular}
    }
    \caption{
    Analysis of setting different numbers of retrieved passages on TopiOCQA.
    }
    \label{table:appendix:retrieved_passages_numbers}
    \vspace{-0.4cm}
\end{table}

\subsection{Analysis of Passage Size in  Retrieval Preference Data Construction} We construct retrieval preference data by retrieving the top-T relevant passages for each rewritten query and measuring their intersections. We tried setting different T values to see their impact on model performance. As shown in the Table~\ref{table:appendix:retrieved_passages_numbers}, the performance is better when T is 50.  Using either a smaller or larger T value will lead to bias. This is attributed to the fact that a smaller T leads to an excessively small intersection between passages retrieved by different queries, with a large portion of intersections being empty. Conversely, a larger T results in an excessively large intersection between passages retrieved by queries, which also weakens the measurement of retrieval consistency scores. However, due to resource constraints and the subtle effects brought by different T values, we did not make more detailed adjustments to the value of T.

\begin{table}[t]
    \small
    \centering
    \resizebox{\linewidth}{!}{
    \begin{tabular}{lccc}
        \toprule
        \multicolumn{1}{c}{\textbf{Method}} &  ROUGE$_1$ & ROUGE$_L$ & BertScore \\        
        \midrule
        LLM4CS	& 29.20	& 27.78	& 85.88 \\
        RETPO	& 26.01	& 24.39	& 85.74 \\
        MSPA-CQR	& 31.97	& 29.85	& 86.72 \\
        \bottomrule
    \end{tabular}
    }
    \caption{
    Performance analysis of end-to-end question answering using the MSPA-CQR and baseline methods.
    }
    \label{table:appendix:end_to_end_qa}
    \vspace{-2mm}
\end{table}

\subsection{End-to-end QA Quality} CQR mainly focuses on the evaluation of retrieval performance, as the performance of retrieval directly affects the effect of response generation. Following previous related research, we only report the retrieval performance.

We evaluated the quality of question answering, using the prompt in Figure~\ref{fig:retrieval_response_pref}. We use LLM4CS and RETPO as baselines for comparison. As shown in Table~\ref{table:appendix:end_to_end_qa}, the quality of responses generated based on the passages retrieved by MSPA-CQR exceeds that of previous methods, consistent with the results observed in the retrieval performance evaluation. This benefits from our design of three-dimensional preferences that can guide the model to generate more semantically comprehensive queries, retrieve more relevant passages, and assist in enhancing the question-answering capability.

\begin{table}[t]
    \small
    \centering
    \resizebox{\linewidth}{!}{
    \begin{tabular}{lcccc}
        \toprule
        & \multicolumn{4}{c}{\textbf{TopiOCQA}} \\
         \cmidrule(lr){2-5}
        \multicolumn{1}{c}{\textbf{Variant}} &  \textbf{MRR} & \textbf{NDCG} & \textbf{R@10}  & \textbf{R@100} \\        
        \midrule
        MSPA-CQR (\texttt{Llama-2-7b-hf}) & 41.4 & 39.5 & 63.5  & 77.4 \\
        MSPA-CQR (\texttt{Llama-3.1-8B-Instruct})	& 40.5	& 38.8	& 62.6	& 76.8 \\
        MSPA-CQR (\texttt{Llama-3.2-3B-Instruct})	& 40.7	& 39.1	& 62.4	& 76.5 \\
        \bottomrule
    \end{tabular}
    }
    \caption{
    Analysis of using different backbones in MSPA-CQR.
    }
    \label{table:appendix:backbones}
    \vspace{-4mm}
\end{table}

\subsection{Other LLMs as Backbones} We conducted experiments using \texttt{Llama-3.1-8B-Instruct} and \texttt{Llama-3.2-3B-Instruct} as backbone models, with the experimental results shown in Table~\ref{table:appendix:backbones}. After using Llama-3.1-8B-Instruct, there is a slight improvement in performance, and the performance of Llama-3.2-3B-Instruct is almost on par with the originally used Llama-2-7b-hf. This indicates that MSPA-CQR can be adapted to different models.

\begin{table}[t]
    \small
    \centering
    \resizebox{\linewidth}{!}{
    \begin{tabular}{lcccccc}
        \toprule
        & \multicolumn{6}{c}{\textbf{TopiOCQA}} \\
         \cmidrule(lr){2-7}
        \multicolumn{1}{c}{\textbf{Variant}} &  \textbf{MRR} & \textbf{NDCG} & \textbf{R@10}  & \textbf{R@100} & LT$_\text{rewrite}$  & LT$_\text{retrieval}$ \\        
        \midrule
        MSPA-CQR	& 30.6	& 29.5	& 51.9	& 75.2	& 1.32	& 0.00079 \\
        MSPA-CQR (merge)	& 29.4	& 28.1	& 50.1	& 74.0	& 0.69 & 0.00034 \\
        RETPO	& 28.3	& 26.5	& 48.3	& 73.1	& 0.57	& 0.00031 \\
        \bottomrule
    \end{tabular}
    }
    \caption{
    Analysis of the trade-off between performance and efficiency.
    }
    \label{table:appendix:latency}
    \vspace{-4mm}
\end{table}

\begin{figure}[t]
    \centering
   \begin{tcolorbox}[
      colback=white, colframe=black, arc=3mm, width=\linewidth,
      title=\textbf{Rewritten Queries Merge}, 
      coltitle=white, colbacktitle=gray, fonttitle=\bfseries
    ]
    $q_1$: Why did Patsy Cline start singing despite the challenges she faced in her personal life?
    
    $q_2$: Why did Patsy Cline start singing?
    
    $q_3$: Why did Patsy Cline start singing, and what were the early influences on her career?
    
    Merged query $q$: What motivated Patsy Cline to start singing and what were her early influences, despite her personal challenges?
    \end{tcolorbox}
    \caption{A case of merging multiple rewritten queries into a single query.}
    \label{fig:rewritten_queries_merge}
\end{figure}

\subsection{Latency Analysis and Solutions}
In MSPA-CQR, we need to generate queries in three dimensions. We found that on TopiOCQA and QReCC, the average time overhead for generating three-dimensional queries per sample is 1.27s and 1.34s respectively, which is acceptable for humans.

To further analyze the trade-off between performance and latency, we comprehensively compared the retrieval performance of rewritten queries, the latency of rewriting, and the latency of retrieval. Taking sparse retrieval on TopiOCQA as an example, we compare the performance of MSPA-CQR and RETPO as well as the latency in rewriting and retrieval in Table~\ref{table:appendix:latency}. Here, LT$_\text{rewrite}$ and LT$_\text{retrieval}$ represent the latency (in seconds) for rewriting and retrieval respectively. Compared to rewrite latency, retrieval latency can be negligible. The rewrite latency of MSPA-CQR only increased by 0.75 compared to RETPO, but we achieved a greater improvement in retrieval performance (an increase of 2.1 in R@100), indicating that MSPA-CQR has effectively balanced performance and efficiency.

We have experimented with another approach to reduce latency. Specifically, we feed the constructed preference query corresponding to the three dimensions of each sample to an LLM and let it merge the three queries to generate a more comprehensive and concise query. We provide such an example in Figure~\ref{fig:rewritten_queries_merge}.

Given the queries $q_1$, $q_2$, and $q_3$, we use an LLM merge them into a single query $q$. In the direct preference optimization phase, we only need to train the generation of the merged query based on the conversation history and the original query. With this approach, in the inference phase, we only need the model to perform a single inference pass. As shown in Table~\ref{table:appendix:latency}, this merge-based approach (MSPA-CQR (merge)) reduces the latency but also reduces the performance of the model due to the loss of some information from the original queries during the merging process. How to balance the efficiency and effectiveness of rewriting is a direction worth exploring in the future.

\begin{table}[t]
\small
\centering
\begin{tabularx}{\linewidth}{X}
\toprule
\textbf{Dialogue Context} \\
Q: What is quoll? \\
A: They  are carnivorous marsupials. \\
Q: Where are they found? \\
A: Australia and New Guinea. \\
Q: Where did the latter get it's name from? \\
A: UNANSWERABLE. \\
Q: What kind of people are seen here? \\
A: The island is presently populated by almost a thousand different tribal groups. \\
... \\
Q: When did the aforementioned department come into action? \\
\midrule
\textbf{Gold Passage}: Department of Environment and Conservation (Western Australia) Introduction The Department of Environment and Conservation (DEC) was a department of the Government of Western Australia that was responsible for implementing the state's conservation and environment legislation and regulations. It was formed on 1 July 2006 by the amalgamation of the Department of Environment and the Department of Conservation and Land Management ... \\
\midrule
\textbf{Ground Truth}: When did the department of environment and conservation (western australia) come into action? (\textbf{Rank: 1})\\
\midrule
\textbf{Rewrite Preferred Query}: When did the aforementioned department, the department of environment and conservation (western australia), come into action to protect the quoll species? (\textbf{Rank: 18})\\
\midrule
\textbf{Retrieval Preferred Query}: When did the aforementioned department come into action? (\textbf{Rank: Not Found})\\
\midrule
\textbf{Response Preferred Query}: When did the department of environment and conservation (western australia) come into action to protect quoll populations in the jarrah forest? (\textbf{Rank: 36})\\
\midrule
\textbf{Rewrite+Retrieval+Response}: \textbf{Rank: 4}\\
\bottomrule
\end{tabularx}
\caption{A case study on TopiOCQA.}
\label{table:case_study1}
\end{table}

\section{Cases of MSPA-CQR}
\label{sec:examples}
In Tables~\ref{table:case_study1},~\ref{table:case_study2}, and~\ref{table:case_study3}, we provide three cases of MSPA-CQR on TopiOCQA and QReCC.

\begin{table}[t]
\small
\centering
\begin{tabularx}{\linewidth}{X}
\toprule
\textbf{Dialogue Context} \\
Q: What happened in 1977 to Journey the band? \\
A: Journey's album sales did not improve and Columbia Records requested that they change their musical style and add a frontman, with whom keyboardist Gregg Rolie could share lead vocal duties. \\
Q: Why did they change their musical direction? \\
A: Journey hired Robert Fleischman and made the transition to a more popular style, akin to that of Foreigner and Boston. \\
Q: What was Robert's role with the band? \\
A: Lead singer. \\
Q: Did they release any new albums? \\
A: Journey released their fourth album, Infinity. \\
... \\
Q: What else did you find interesting? \\
\midrule
\textbf{Gold Passage}: United Kingdom in the Eurovision Song Contest - Wikipedia CentralNotice United Kingdom in the Eurovision Song Contest From Wikipedia, the free encyclopedia Jump to navigation Jump to search United Kingdom Member station BBC National selection events National Final Festival of British Popular Songs 1957 A Song for Europe 1959–1963 1964–1975 (song) ... \\
\midrule
\textbf{Ground Truth}: What else did you find interesting after Journey set on a new path with a more mainstream sound to make their highest chart success to date? (\textbf{Rank: 28})\\
\midrule
\textbf{Rewrite Preferred Query}: What else did you find interesting about Journey the band during their time in the 1970s and 1980s? (\textbf{Rank: Not Found})\\
\midrule
\textbf{Retrieval Preferred Query}: What else did you find interesting about the band Journey besides the release of their fourth album, Infinity? (\textbf{Rank: Not Found})\\
\midrule
\textbf{Response Preferred Query}: What else did you find interesting about Journey the band's 1977-1981 period? (\textbf{Rank: Not Found})\\
\midrule
\textbf{Rewrite+Retrieval+Response}: \textbf{Rank: 2}\\
\bottomrule
\end{tabularx}
\caption{A case study on QReCC.}
\label{table:case_study2}
\end{table}

\begin{table}[t]
\small
\centering
\begin{tabularx}{\linewidth}{X}
\toprule
\textbf{Dialogue Context} \\
Q: Where did the uk come in eurovision song contest? \\
A: The United Kingdom came in 26th place in the Eurovision Song Contest 2019. \\
Q: What song did the country perform? \\
A: The United Kingdom's selected song for the Eurovision Song Contest 2019 was Bigger than Us by Michael Rice. \\
Q: How many times has the uk participated in the contest? \\
A: The United Kingdom has participated in the Eurovision Song Contest 62 times. \\
Q: What other countries are part of the big 5? \\
\midrule
\textbf{Gold Passage}: United Kingdom in the Eurovision Song Contest - Wikipedia CentralNotice United Kingdom in the Eurovision Song Contest From Wikipedia, the free encyclopedia Jump to navigation Jump to search United Kingdom Member station BBC National selection events National Final Festival of British Popular Songs 1957 A Song for Europe 1959–1963 1964–1975 (song) ... \\
\midrule
\textbf{Ground Truth}: What other countries are part of the eurovision song contest big 5 besides the uk? (\textbf{Rank: 2})\\
\midrule
\textbf{Rewrite Preferred Query}: What other countries are part of the Big 5 in the Eurovision Song Contest? (\textbf{Rank: 13})\\
\midrule
\textbf{Retrieval Preferred Query}: What other countries are part of the big 5 besides the uk? (\textbf{Rank: Not Found})\\
\midrule
\textbf{Response Preferred Query}: What other countries are part of the eurovision big 5 besides the uk? (\textbf{Rank: 16})\\
\midrule
\textbf{Rewrite+Retrieval+Response}: \textbf{Rank: 4}\\
\bottomrule
\end{tabularx}
\caption{A case study on QReCC.}
\label{table:case_study3}
\end{table}

\end{document}